\pdfoutput=1

\documentclass[11pt]{article}

\usepackage{ACL2023}

\usepackage{times}
\usepackage{latexsym}

\usepackage[T1]{fontenc}

\usepackage[utf8]{inputenc}

\usepackage{microtype}

\usepackage{inconsolata}

\newcommand{\model}{\textsc{glimmer}\xspace}
\newcommand{\glimmer}{\textsc{glimmer}\xspace}

\newcommand{\lumen}{\textsc{lumen}\xspace}
\newcommand{\fid}{FiD\xspace}
\newcommand{\papertitle}{\model: generalized late-interaction memory reranker}

\newcommand{\liveprop}{\alpha}
\newcommand{\rerankprop}{\beta}
\definecolor{fidcolor}{rgb}{0.9296875, 0.3984375, 0.46484375} 
\definecolor{lumencolor}{rgb}{0.95, 0.52, 0.0} 
\definecolor{glimmercolor}{rgb}{0, 0.46484375, 0.73046875} 
\definecolor{darkgrey}{rgb}{0.33203125,0.33203125,0.33203125} 

\definecolor{flycolor}{rgb}{0, 0.46484375, 0.73046875} 
\definecolor{precolor}{rgb}{0.95, 0.52, 0.0} 

\definecolor{singlecolor}{rgb}{0.85, 0.75, 0.85} 
\definecolor{multicolor}{rgb}{0.18, 0.55, 0.34} 

\newcommand{\modelmark}{square*}

\newcommand{\propaxis}{Live proportion $\liveprop$}

\usepackage{times}
\usepackage{latexsym}
\usepackage{xspace}
\usepackage{amsmath}
\usepackage{tabularx}
\usepackage{multirow}
\usepackage{booktabs}

\usepackage{xcolor}
\definecolor{dgreen}{rgb}{0,0.5,0}

\usepackage{tikz}
\pgfdeclarelayer{background}
\pgfdeclarelayer{main}
\pgfsetlayers{background,main}
\usetikzlibrary{calc}
\usepackage{ifthen}

\usepackage{graphicx}
\usepackage{caption}
\usepackage{subcaption}
\usepackage{float}
\usepackage{stfloats}
\usepackage{tikz}
\usepackage{nicefrac}
\usepackage{pgfplots}
\usepgfplotslibrary{fillbetween}
\usetikzlibrary{patterns}
\pgfplotsset{compat=1.8}

\newenvironment{customlegend}[1][]{%
    \begingroup
    \csname pgfplots@init@cleared@structures\endcsname
    \pgfplotsset{#1}%
}{%
    \csname pgfplots@createlegend\endcsname
    \endgroup
}%

\def\addlegendimage{\csname pgfplots@addlegendimage\endcsname}

\title{\papertitle}

\author{
Michiel de Jong\thanks{\- \ Equal contribution.} \ \footnotemark[2]\thanks{\- \ University of Southern California. Work done at Google Research.} ,~ Yury Zemlyanskiy\footnotemark[1] \\   {\bf Nicholas FitzGerald},~ {\bf Fei Sha},~ {\bf Sumit Sanghai},~ {\bf William W. Cohen}, ~ {\bf Joshua Ainslie}
  \AND
  {\rm \Large Google Research}\\
  }

\begin{document}
\maketitle
\begin{abstract}
Memory augmentation is a powerful approach for efficiently incorporating external information into language models, but leads to reduced performance relative to retrieving text. Recent work introduced \lumen, a memory-retrieval hybrid that partially pre-computes memory and updates memory representations on the fly with a smaller live encoder.

We propose \model, which improves on this approach through 1) exploiting free access to the powerful memory representations by applying a shallow reranker on top of memory to drastically improve retrieval quality at low cost, and 2) incorporating multi-task training to learn a general and higher quality memory and live encoder. \model achieves strong gains in performance at faster speeds compared to \lumen and \fid on the KILT benchmark of knowledge-intensive tasks.
\end{abstract}

\section{Introduction}
\label{section:intro}

Retrieval-augmented language models achieve strong performance, but are computationally expensive due to the need to process retrieved passages. A large body of work attempts to reduce the cost of reading retrieved passages through conditional computation~\citep{colt5, canext, schuster2022confident}, reranking~\citep{r3rerank, kgfid, r3rerank}, or memory~\citep{tome, memorizing, fidmemory}.

Reranking improves retrieval quality and therefore reduces the number of passages that need to be processed by the reader. However, neural reranking is expensive, as each retrieved candidate is processed by a neural network. Late interaction rerankers~\citep{colbert, sdr, prettr} pre-compute intermediate token representations and apply a smaller neural model on the fly to combine query and document representations and produce a ranking score. Late interaction drastically improves speed at the cost of storage and pre-computation overhead and machinery.

Recently the idea of late-interaction has also been applied to retrieval augmented generation: \lumen~\citep{lumen} interpolates between memory and retrieval augmentation to achieve a better quality-compute trade-off.

We propose \model (Generalized Late-Interaction Memory Reranker), a late interaction approach that combines these lines of work by \textit{unifying reranking and memory into a single end-to-end model}. Like \lumen, \model consists of a memory encoder that generates pre-computed token representations for retrieval documents, and a live encoder that combines the representations of retrieved documents with the query. After the first layers of the live-encoder, a ranking layer selects the most relevant passages which are retained for further processing. The model is trained to rank passages by usefulness to the reader through a perplexity distillation auxiliary loss~\citep{atlas}.

\model also improves on \lumen by using a single general memory and live encoder over all tasks, trained with multi-task fine-tuning over knowledge intensive datasets. 

We evaluate on the KILT benchmark of knowledge-intensive tasks~\citep{kilt}. We first find that multi-task training of the memory and live encoders strongly improves model quality relative to training on a single task, especially when devoting less capacity to the live encoder. Moreover, \model strongly improves over both multi-task trained \lumen and FiD in both quality and speed. In general, \model successfully unifies reranking and memory into a single efficient, high-quality model.

\begin{figure*}[t!]
    \centering
    \includegraphics[width=0.99\linewidth]{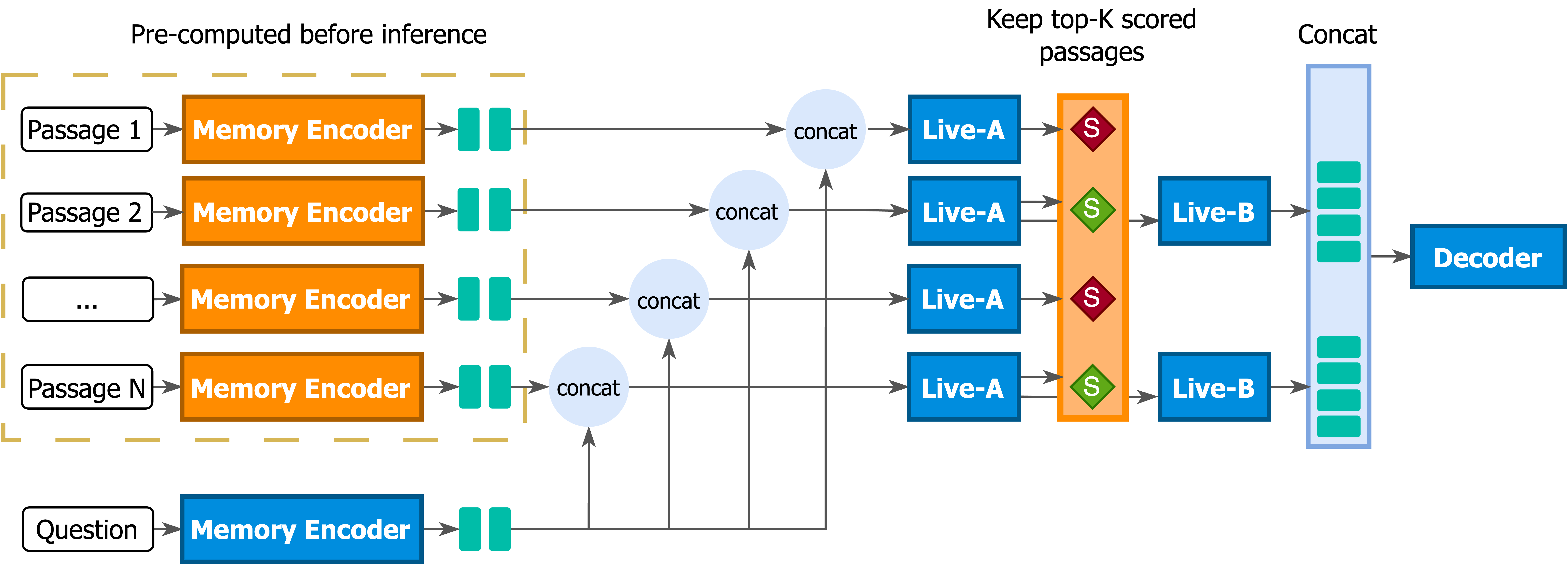}
    \caption{Overview of \model architecture. \\{\bf Memory:} The memory encoder is updated during multi-task training, unlike \lumen, before being applied to the corpus to generate partially pre-computed memory representations. The memory encoder is also applied during inference to generate partial question representations that are compatible with the memory. \\{\bf Live:} Each passage memory is concatenated with the question representation, and a live encoder (proportion $\liveprop$ of the total model) is then applied to condition the passage on the input in two stages. After the first stage, consisting of a fraction $\rerankprop$ of live layers, a scoring layer selects a small subset of high-scoring relevant passages to keep and less relevant passages are discarded. The selected passage representations are updated by the second stage of the live encoder. Finally, the conditioned representations are concatenated and attended to by the decoder as in \fid.}
    \label{fig:architecture}
\end{figure*}

\section{Background}
\label{section:background}

We are interested in achieving the best possible trade-off between quality and inference compute. The following section describes \fid and \lumen, the baseline methods that \glimmer is built on, and their computational properties. A more in-depth analysis of these methods can be found in \citet{lumen}.

\subsection{Fusion-in-Decoder}

Fusion-in-Decoder \citep{fid} is based on a T5 encoder-decoder model~\citep{t5}. For each input, a number of relevant text passages are retrieved, and the input is prepended to each passage. The resulting input-passage pairs are encoded separately by the encoder, and the encoded pairs are then concatenated into a flat sequence of token representations and attended to by the decoder to produce a target output. For each model, \textcolor{flycolor}{\textbf{live}} components are in blue and components \textcolor{precolor}{\textbf{pre-computed}} before inference in orange.
\begin{equation*}
    G = \text{\textbf{\textcolor{flycolor}{Dec}}}\Big[\text{\textcolor{flycolor}{\textbf{Enc}}}(Q; \text{Passage}_1); \ldots \text{\textbf{\textcolor{flycolor}{Enc}}}(Q; \text{Passage}_k)\Big]
\end{equation*}
Let $k$ be the number of passages, $n_p$ be the number of tokens per passage, $n_t$ the number of target tokens, $L$ the number of layers, and $d$ the dimension of the model. Following analysis from \citet{dejong2022fido, lumen}, the FLOPs for a single inference sample of FiD (ignoring attention score computation) is given by
\begin{equation*}
    F_{\fid} = \underbrace{k n_p \cdot L \cdot 14 d^2}_{\text{Encoder and cross-attention}} + \underbrace{n_t \cdot L \cdot 14 d^2}_{\text{Decoder}}
\end{equation*}
with factors $8d^2$ per token from feedforward layers, $4d^2$ from self-attention projection layers, and $2d^2$ from cross-attention projection layers. \citet{lumen} contains a derivation of \fid model complexity in greater detail.

\subsection{\lumen}

Typically the combined length of retrieved passages is much larger than the target length, such that the majority of FLOPs are consumed by the encoder processing retrieved passages. \lumen reduces encoder inference cost by partially pre-computing the encoder representation for retrieved passages. At inference time, \lumen retrieves the intermediate layer representations rather than the text.

More precisely, \lumen is initialized from a pre-trained T5 encoder-decoder model. The decoder functions the same as the standard FiD decoder, but the T5 encoder is divided into a large \textcolor{precolor}{\textbf{memory encoder}} which contains the first $1 - \liveprop$ proportion of layers, and a smaller \textcolor{flycolor}{\textbf{live encoder}} with the remaining $\liveprop$ proportion of layers. The memory encoder is applied offline to passages in the corpus to pre-compute memory representations, which are later updated conditioned on input and task on the fly by the fine-tuned live encoder. In order to ensure that memory representations and input are compatible, \lumen applies the memory encoder\footnote{The original \lumen implementation used a separate question encoder, but we show this is unnecessary.} to the input before prepending the question representation to the memory representation.
\vspace{-3pt}
\begin{align*}
    & H_i = \Big[\textbf{\textcolor{flycolor}{MemEnc}}(Q);\hspace{0.2cm} \text{\textbf{\textcolor{precolor}{MemEnc}}}(\text{Passage}_i)\Big] \\
    &G = \text{\textbf{\textcolor{flycolor}{Dec}}}\Big[Q; \text{\textbf{\textcolor{flycolor}{LiveEnc}}}(H_1); \ldots \text{\textbf{\textcolor{flycolor}{LiveEnc}}}(H_k) \Big]
\end{align*} 
Choosing $\liveprop = 1$ yields a model very close to FiD while $\liveprop = 0$ is a full memory model.
During inference \lumen applies only a proportion $\liveprop$ of the layers, leading to a fraction $\liveprop$ of FiD reader FLOPs for any given model size.
\begin{align*}
    F_{\lumen} &= \underbrace{kn_p \cdot \liveprop L \cdot 12d^2}_{\text{Encoder}} \\
    &+ \underbrace{k n_p \cdot L \cdot 2d^2}_{\text{Cross-attention}}
    + \underbrace{n_t \cdot L \cdot 14 d^2}_{\text{Decoder}}
\end{align*}
\section{\model}
\label{section:method}

\model builds on \lumen with two major differences: \model incorporates a built-in reranker, and shares the memory and live encoder across many tasks. Standard reranking approaches struggle with a trade-off: smaller models may not be sufficiently powerful to judge whether a passage is relevant to an input, while the cost of larger models defeats a large part of the purpose of using a reranker in the first place. The \lumen architecture offers an opportunity to circumvent this trade-off, as the majority of the passage representations are pre-computed. \model re-uses the initial layers of the live encoder for reranking, yielding a powerful re-ranking model at relatively modest computational cost. 

Sharing weights across tasks, meanwhile, allows for training the memory encoder without storing duplicate pre-computed representations, and strongly increases the effectiveness of the live encoder. Figure \ref{fig:architecture} shows an overview of the \model architecture.

\subsection{Architecture}
Compared to \lumen, \model divides the live encoder into two components, where the first component is responsible for initial interaction and reranking and the second component performs further processing on representations of selected passages. The first component contains $\rerankprop$ proportion of live encoder layers with the remainder of layers in the second component. After the first live encoder, a linear projection layer is applied to the first token of each input-passage pair to generate a relevance score for the passage. The top-$m$ passages with the highest scores out of the original $k$ are processed by the second live encoder, and the other passages are discarded. The output of the second live encoder is fed to the decoder as in \fid and \lumen.
\begin{align*}
\label{eqn:glimmer}
    &H_i = \Big[\textbf{\textcolor{flycolor}{MemEnc}}(Q);\hspace{0.2cm} \text{\textbf{\textcolor{precolor}{MemEnc}}}(\text{Passage}_i)\Big]\\
    &H'_i = \text{\textbf{\textcolor{flycolor}{LiveEncA}}}(H_i)\\
    &R_j = H'_i \ \ \text{s.t. Rank} \ [\textbf{\textcolor{flycolor}{Score}}(H'_i)] = j \\    
    &G = \text{\textbf{\textcolor{flycolor}{Dec}}}\Big[Q; \text{\textbf{\textcolor{flycolor}{LiveEncB}}}(R_1); \ldots \text{\textbf{\textcolor{flycolor}{LiveEncB}}}(R_m) \Big] \\
\end{align*} 
\subsection{Training}

The memory encoder, both live encoder components, the scoring projection and the decoder are all trained end-to-end. Unlike in \lumen, the memory encoder does not need to be frozen as we share a single memory encoder between all tasks. In order to train the scoring projection and encourage the memory and first live encoder to produce representations suitable for reranking, we employ an auxiliary perplexity distillation loss~\citep{atlas}. This loss encourages the model to rank passages by how much they lower the perplexity of the final generation, if that input-passage was fed to the decoder by itself. In particular, perplexity distillation minimizes the KL-divergence between the distribution implied by the reranking scores (computed from the output of the first live encoder component applied to concatenation of input and passage representations) and the distribution implied by the resulting perplexities:
\begin{equation*}
    p^{\text{rank}}_k = \frac{\exp(\text{Score}(\text{Passage}_k, Q) / \tau)}{\sum_i \exp(\text{Score}(\text{Passage,}_i, Q) / \tau)}
\end{equation*}
\begin{equation*}
    p^{\text{LM}}_k = \frac{\exp(\log p_{LM}(\text{Answer} | \text{Passage}_k, Q) / \tau)}{\sum_i \exp(\log p_{LM}(\text{Answer} | \text{Passage}_i, Q) / \tau)}
\end{equation*}

\begin{equation*}
    \mathcal{L}_{\text{pdist}} = KL(p^{\text{rank}}, \  p^{\text{LM}})
\end{equation*}

\subsection{Computational analysis}

The difference in computational complexity between \model and \lumen lies in reranking. The $m$ selected passages are processed by the entire live encoder and then fed through the decoder, yielding computational cost equal to applying \lumen with $m$ passages (less than the full number of retrieved passages $k$). However, for the passages that were not selected, \model still applied the first live encoder component, leading to a reranking cost:
\begin{equation*}
    F_{\model} = F_{\lumen}^m + \underbrace{(k - m)n_p \cdot \rerankprop \liveprop L \cdot 12d^2}_{\text{Reranking}} 
\end{equation*}

If we use a small number of selected passages $m << k$ and small fraction of reranking layers $\rerankprop << 1$, then \model is significantly less computationally intensive than \lumen.  
\begin{figure*}[ht!]
    \centering
    \begin{minipage}[c]{0.45\textwidth}
    \begin{subfigure}[t]{\linewidth}
        \begin{tikzpicture}
            \begin{axis}[
                xbar,
                bar shift=0pt,
                y dir=reverse,
                xlabel={Performance},
                width=0.99\columnwidth,
                height=0.56\columnwidth,
                symbolic y coords={FiD, LUMEN, GLIMMER},
                yticklabels={\small{FiD}, \small{LUMEN}, \small{GLIMMER}},   
                ytick={FiD, LUMEN, GLIMMER},
                ytick style={draw=none},
                xmin=68, xmax=73.8,
                enlarge y limits=0.4,
                xmajorgrids=true,
                grid style=dashed,
            ]
            \addplot[fill=fidcolor] coordinates {(69.215,FiD)};
            \addplot[fill=lumencolor] coordinates {(70.98166667,LUMEN)};
            \addplot[fill=glimmercolor] coordinates {(73.22,GLIMMER)};            
            \end{axis}
        \end{tikzpicture}
        \label{fig:bar_chart}
    \end{subfigure}
      \vspace{1cm}
    \begin{subfigure}[t]{\linewidth}
        \begin{tikzpicture}
            \begin{axis}[
                xbar,
                bar shift=0pt,
                y dir=reverse,
                xlabel={Samples per TFLOP},
                width=0.99\columnwidth,
                height=0.56\columnwidth,
                symbolic y coords={FiD, LUMEN, GLIMMER},
                yticklabels={\small{FiD}, \small{LUMEN}, \small{GLIMMER}},   
                ytick={FiD, LUMEN, GLIMMER},
                ytick style={draw=none},                
                xticklabel style={
                    /pgf/number format/fixed,
                    /pgf/number format/precision=2
                },                                
                xmin=0.08, xmax=0.14,
                enlarge y limits=0.4,
                xmajorgrids=true,
                grid style=dashed,
            ]
            \addplot[fill=fidcolor] coordinates {(0.09272427067,FiD)};
            \addplot[fill=lumencolor] coordinates {(0.1067540778,LUMEN)};
            \addplot[fill=glimmercolor] coordinates {(0.1278586731,GLIMMER)};            
            \end{axis}
        \end{tikzpicture}
        \label{fig:bar_chart}
    \end{subfigure}
    \end{minipage}    
    \hfill
    \begin{minipage}[c]{0.45\textwidth}
    \vspace{-1.5cm}
    \begin{tikzpicture}
    \begin{axis}[
    width=0.98\columnwidth,
    height=1.1\columnwidth,
    xmin=0.08, xmax=0.14,
    ymin=67.8, ymax=74.0,
    ylabel={Performance},
    xlabel={Samples per TFLOP},
    xticklabel style={
        /pgf/number format/fixed,
        /pgf/number format/precision=2
    },                
    legend columns=1,
    legend cell align=left,
    legend style={
        anchor=south,
        at={(0.84, 0.05)},
    },
    ]
    \addplot[only marks, mark=*, mark options={draw=fidcolor, fill=fidcolor, scale=2}] coordinates {
        (0.09272427067,69.215)
    };
    \node at (axis cs:0.089,69.00) [anchor= north west, color=black] {\small{\fid}};      
    
    \addplot[only marks, mark=*, mark options={draw=lumencolor, fill=lumencolor, scale=2}] coordinates {
        (0.1067540778,70.98166667)
    };
    \node at (axis cs:0.10,70.7) [anchor= north west, color=black] {\small{\lumen}};
    
    \addplot[only marks, mark=*, mark options={draw=glimmercolor, fill=glimmercolor, scale=2}] coordinates {
        (0.1278586731,73.22)
    };
    \node at (axis cs:0.120,72.95) [anchor= north west, color=black] {\small{\model}};
    \end{axis}
    \end{tikzpicture}    
    \end{minipage}    
    \vspace*{-10mm}
    \caption{\textbf{\model is faster and higher quality than \lumen which in turn is faster and higher quality than \fid.} Comparison of \model, \lumen and \fid XXL model average performance on KILT dev set, and inference speed. \fid uses 5 retrieved passages, \lumen uses 10 retrieved passages, and \model uses 25 retrieved passages, reranked to 5 final passages. \lumen and \model have live proportion $\alpha = \frac{1}{3}$.}
    \label{fig:main_results}
\end{figure*}
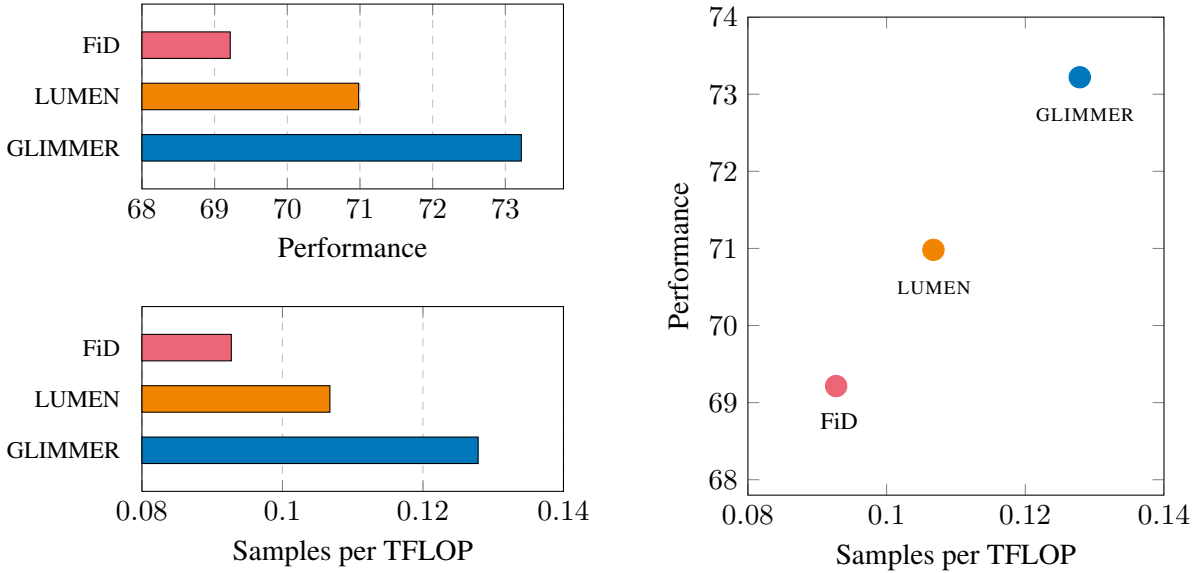 with k retrievals.

We note that this computational analysis is limited to FLOPs, rather than practical latency. For autoregressive inference, the decoder is often bottlenecked by memory bandwidth rather than FLOPs~\cite{shazeer2019mq, dejong2022fido}. However, many recent techniques ameliorate this constraint, such as flavors of multi-query attention~\citep{shazeer2019mq, gqa}, layer sparsity~\citep{dejong2022fido}, speculative decoding~\citep{specleviathan, specchen}, and others. Any model deployed in an environment where inference speed is important will likely employ one or more such techniques, such that FLOPs are a binding constraint. For the rest of this paper, we will measure computational cost in FLOPs; \citet{lumen} contains analysis for how FLOPs and latency interact for \lumen.

As we will show, \model represents a better quality-compute trade-off than \lumen and \fid. 

\section{Experiments}
\label{section:experiments}

\subsection{Experimental setup}

\paragraph{Model configuration}
\model is based on the T5.1.1 architecture~\citep{t5} like \lumen, implemented in JAX~\citep{flax}, Flax~\citep{flax} and Flaxformer. All models are initialized from public T5.1.1 checkpoints. FiD is fine-tuned according to the recipe from the original paper~\citep{fid}. For \lumen and \model, given proportion of live layers $\liveprop$, the memory encoder is initialized with the first 1 - $\liveprop$ proportion of layers of the T5 encoder, and the live encoder is initialized with the last $\liveprop$ proportion of layers of the T5 encoder. Main experiments use $\liveprop = \frac{1}{3}$.

\paragraph{Fine-tuning}
For fine-tuning we use the Adafactor optimizer~\citep{adafactor} with constant learning rate of 0.0001, batch size 128, and dropout rate 0.1 for all tasks. For multi-task training we sample uniformly from tasks. We allocate 48 tokens for the question and 304 tokens for each passage. In addition to the standard language modeling loss, reranking experiments use an auxiliary perplexity distillation loss with weight and temperature 1.0. We train until convergence and select the checkpoint with the highest performance on the dev set. We use greedy decoding for inference.

\paragraph{Data}
We train and evaluate on a subset of datasets from the KILT benchmark of knowledge-intensive tasks~\citep{kilt}. In particular, this includes question answering datasets Natural Questions~\citep{nq}, TriviaQA~\citep{triviaqa}, and HotPotQA~\citep{hotpotqa}, fact verification dataset FEVER~\citep{fever}, and slot-filling datasets Zero Shot RE~\citep{zeroshot} and T-REx~\citep{trex}. We apply the relevance filtering procedure from \citet{multikilt} to ameliorate problems from imbalanced datasets.

\paragraph{Retrieval}

We employ the retrieval procedure from \citet{multikilt}. Wikipedia is divided into chunks up to 200 words, and we retrieve the passages with the highest similarity score to the query, computed by a pre-trained GTR-Base model~\citep{gtr}.

\subsection{Main results}
\label{section:main_results}

For our main results, we compare FiD, \lumen (with updated architecture and multi-task training) and \model. Due to in-built reranking, \model processes passages more efficiently and can therefore retrieve more documents than \lumen, which in turn can retrieve more documents than \fid. As Figure \ref{fig:main_results} shows, this efficiency translates into a higher quality and faster model, with \glimmer outperforming \lumen and \fid at faster speed.

\subsection{Retrieval and reranking}
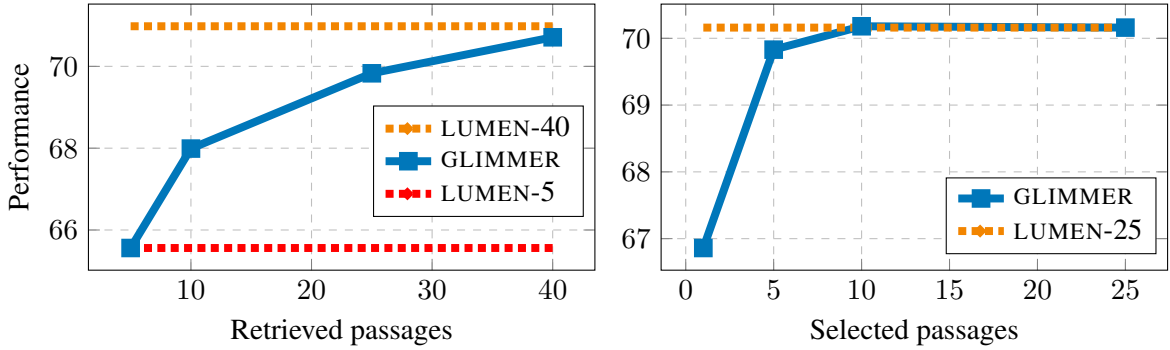
\begin{figure*}
\centering
\begin{subfigure}[t]{0.49\textwidth}
    \begin{tikzpicture}
            \begin{axis}[
            scale only axis,
            width=0.85\columnwidth,
            height=0.45\columnwidth,
            ylabel={Performance},
            xlabel={Retrieved passages},
            mark=x,
            ymajorgrids=true,
            xmajorgrids=true,
            xminorticks=true,
            grid style=dashed,
            legend columns=1,
            legend cell align=left,
            legend style={
                anchor=south,
                at={(0.77, 0.2)},
            },
        ]
      
            \addplot[color=lumencolor,line width=3, dotted] table {
                5 70.98  
            	40 70.98
            };
            \addplot[color=glimmercolor,mark=\modelmark,mark size=2pt,line width=3] table {
                5 65.56
                10 67.99
                25 69.83
                40 70.71
            };
            \addplot[color=red,line width=3, dotted] table {
                5 65.56  
            	40 65.56
            };                 
            
            \legend{\lumen-40, \glimmer, \lumen-5}           
            \end{axis} 
    \end{tikzpicture}
     \end{subfigure}
     \hfill
     \begin{subfigure}[t]{0.49\textwidth}
    \begin{tikzpicture}
            \begin{axis}[
            scale only axis,
            width=0.85\columnwidth,
            height=0.45\columnwidth,
            xlabel={Selected passages},
            mark=x,
            ymajorgrids=true,
            xmajorgrids=true,
            xminorticks=true,
            grid style=dashed,
            legend columns=1,
            legend cell align=left,
            legend style={
                anchor=south,
                at={(0.77, 0.06)},
            },
        ]
            \addplot[color=glimmercolor,mark=\modelmark,mark size=2pt,line width=3] table {
                1 66.86  
                5 69.83
                10 70.18
                25 70.16
            };.16
            \addplot[color=lumencolor,line width=3, dotted] table {
                1 70.16  
            	25 70.16
            };                    
            \legend{\glimmer, \lumen-25}
            \end{axis}   
    \end{tikzpicture}     
     \end{subfigure}
     \hfill
     
\caption{Average dev performance on KILT for \model-Large with live proportion $\frac13$ and rerank proportion $\frac14$ as a function of number of retrievals with 5 selected passages (left) and number of selected passages with 25 retrievals (right).}
\label{fig:num_retrievals}
\end{figure*}
The main results indicate that \model can achieve higher quality at lower cost than \fid and \lumen by retrieving more passages initially and reranking to a much smaller number of passages. Here we investigate how different choices regarding retrieval and reranking affect the results.

\paragraph{Number of retrieved and selected passages} Figure \ref{fig:num_retrievals} shows how performance varies with the total number of retrieved passages and the number of selected passages after reranking. Performance strongly increases in the total number of retrieved passages, with sharply diminishing returns in the number of \emph{selected} passages. These results indicate that the reranker effectively selects useful passages, such that the bottleneck is whether or not the relevant information is present in original retrieved passages.

\begin{figure}[h!]
\begin{tikzpicture}
            \begin{axis}[
            scale only axis,
            width=0.85\columnwidth,
            height=0.45\columnwidth,
            ylabel={Performance},
            xlabel={Rerank proportion $\rerankprop$},
            mark=x,
            ymajorgrids=true,
            xmajorgrids=true,
            xminorticks=true,
            grid style=dashed,
            legend columns=1,
            legend cell align=left,
            legend style={
                anchor=south,
                at={(0.77, 0.06)},
            },
        ]
     
            \addplot[color=glimmercolor,mark=\modelmark,mark size=2pt,line width=3] table {
                0.0 64.53  
                0.125 67.6
                0.25 69.83
                0.5 70.01
                1.0 69.94
            };
            \addplot[color=lumencolor,line width=3, dotted] table {
                0 70.16  
            	1 70.16
            };                       
            \legend{\glimmer, \lumen-25}            
            \end{axis}   
\end{tikzpicture}
    \caption{Average dev performance on KILT for \model-Large with live proportion $\frac13$, 25 retrieved passages and 5 selected passages as a function of rerank proportion $\rerankprop$. Baseline $\rerankprop$ is 0.25, equivalent to 2 reranking layers out of 8 total live layers.}
    \label{fig:num_rerank_layers}
\end{figure}
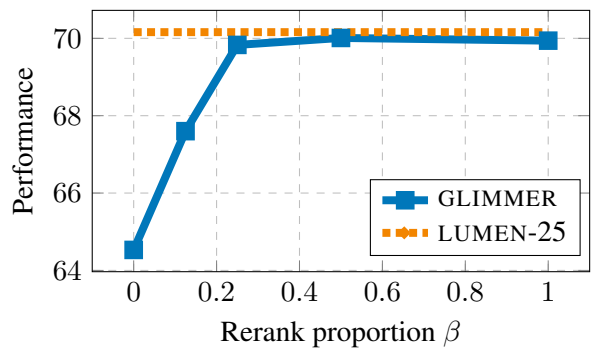
The former intuition is further supported by Figure \ref{fig:num_rerank_layers}, as applying sufficient reranking layers almost recovers the performance of using all 25 retrievals. On the other hand, some neural reranking with full interaction is clearly helpful, as using rerank proportion fewer than 0.25 (fewer than 2 reranking layers) strongly harms performance.

\begin{figure}[h!]
     \centering
    \hspace{-10pt}     
        \begin{tikzpicture}[scale=1.0]
            \begin{axis}[
            scale only axis,
            width=0.82\columnwidth,
            height=0.42\columnwidth,         
            ylabel={Performance},
            xlabel={Selected passages},
            mark=x,
            ymajorgrids=true,
            xmajorgrids=true,
            xminorticks=true,
            grid style=dashed,
            legend columns=1,
            legend cell align=left,
            legend pos={south east},
            ]
            \legend{2 rerank layers, 4 rerank layers}        
            \addplot[color=multicolor,mark=square,mark size=1pt,line width=2] table {
                5 70.540702
                10 70.885407
                15 70.873928
                20 70.95
                40 70.820102
                };            
            \addplot[color=singlecolor,mark=square,mark size=1pt,line width=2] table {
                5 70.851149
                10 71.1
                15 71.083193
                20 70.964394
                40 70.974273
            };
            \end{axis}
        \end{tikzpicture}  
    \caption{Average dev performance on KILT for \model-Large with live proportion $\frac13$ with 40 retrievals as a function of number of selected passages.}
    \label{fig:num_selections}
\end{figure}
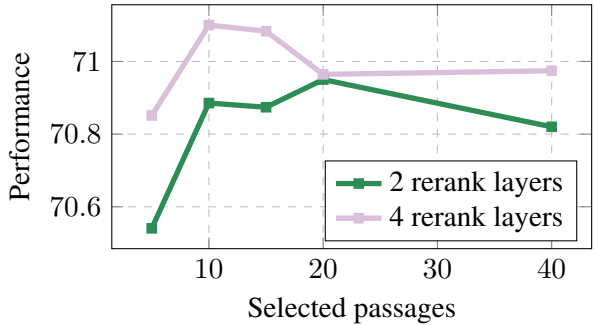

Interestingly, as shown in Figure \ref{fig:num_selections}, with a large number of retrievals, selection is sufficiently accurate that selecting more passages harms performance due to distraction from irrelevant context. The optimal number of selected passages is lower with more reranking layers, as the top ranked passages better capture all useful information.

\begin{table}[ht!]
\centering
\begin{tabular}{lc}
    \textbf{Reranker} & \textbf{Performance} \\
    \toprule
     \model (shared) & 69.8 \\
     Separate (from T5) & 70.0 \\
     Separate (from scratch) & 68.7 \\
    \bottomrule
\end{tabular}
\caption{Average performance on KILT dev sets for \model-Large with 25 retrieved and 5 selected passages for different configurations of the reranker: shared, separately initialized from T5, and separately initialized from scratch.}
\label{table:rerank_share}
\end{table}
\paragraph{Separate reranker} It is also informative to consider the effect of using the live encoder to perform the reranking, as opposed to a separate reranker. Table \ref{table:rerank_share} compares performance of \model with using a separate reranker, initialized from T5 or trained from scratch. We note that using a separate reranker achieves comparable performance at the cost of a more complicated model, and additional memory and computation overhead. Initializing the reranker from pre-trained weights is important - attempting to learn reranking layers from scratch significantly lowers performance.

\subsection{Multi-task training}

The second major improvement in \model is sharing the memory and live encoder between tasks, and consequently training the memory encoder. We present experiments that attempt to disentangle the effects of these improvements.

Figure \ref{fig:multi_vs_single} demonstrates the effect of multi-task training by comparing performance on NQ between models trained only on NQ and models trained on KILT. To isolate the effect of multi-task training, we compare \fid and \lumen, and train the memory for all models in this comparison. Multi-task training significantly benefits all models, but is disproportionately impactful for \lumen, especially with lower live proportions. Figure \ref{fig:perf_vs_prop_for_nq_and_multitask} shows the difference between single and multi-task training as a function of live proportion, with multi-task performance leveling out earlier, further showing larger impact for smaller live proportion.

The late interaction that the live encoder is responsible for is rather different from its pre-training task, so it is intuitive that the live encoder would disproportionately benefit from increased size and diversity of data.
 
Multi-task training also enables learning a memory encoder. Table \ref{table:memory_ablation} shows that training the memory encoder is important for performance, which is expected as the pre-trained encoder is not designed to function as a memory encoder out of the box.

\begin{figure}[h!]  
\centering \hspace{-25pt}
\begin{tikzpicture}[scale=1.0]
\begin{axis}[
    xbar stacked,
    bar width=14pt,
    enlarge y limits=0.25,
    width=\columnwidth,
    height=0.7\columnwidth,
    major y tick style = transparent,    
    xmajorgrids = true,
    xlabel = {Exact match},
    symbolic y coords={L$\nicefrac{1}{8}$, L$\nicefrac{1}{3}$, \fid},
    ytick = data,
    xmin=45,
    axis y line*=none,
    axis x line*=bottom,
]
    \addplot[style={singlecolor,fill=singlecolor,mark=none}]
        coordinates {(51.18,L$\nicefrac{1}{8}$) (58.34,L$\nicefrac{1}{3}$) (59.11,\fid)};
    \addplot[style={multicolor,fill=multicolor,mark=none}]
        coordinates {(6.8,L$\nicefrac{1}{8}$) (2.68,L$\nicefrac{1}{3}$) (1.69,\fid)};        
\end{axis}

\begin{customlegend}[
    legend columns=2,
    legend cell align=left,
    legend style={
        anchor=north east, 
        align=left,
        at={(5, 4.4)},
        column sep=1ex
    },
    legend entries={NQ only, Multi-task}
]
    \addlegendimage{mark=square*,only marks,solid,color=singlecolor}
    \addlegendimage{mark=square*,only marks,solid,color=multicolor}

\end{customlegend}
\end{tikzpicture}            

\caption{{\bf Multi-task training disproportionately benefits \lumen relative to \fid.} Exact match on Natural Questions dev set when trained only on Natural Questions vs on set of KILT tasks for \fid, \model-$\frac13$ and \model-$\frac18$ Large models.}
\label{fig:multi_vs_single}
\end{figure}
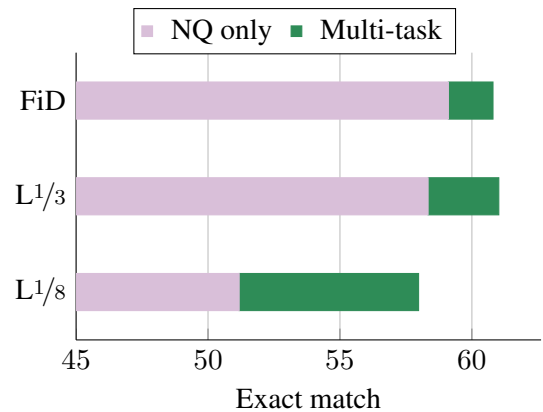
\begin{figure}[h!]
     \centering
    \hspace{-5pt}     
        \begin{tikzpicture}[scale=1.0]
            \begin{axis}[
            scale only axis,
            width=0.85\columnwidth,
            height=0.42\columnwidth,           
            ylabel={Exact Match},
            xlabel={\propaxis},
            mark=x,
            ymajorgrids=true,
            xmajorgrids=true,
            xminorticks=true,
            grid style=dashed,
            legend columns=1,
            legend cell align=left,
            legend pos={south east},
            ]
            \legend{KILT, NQ-only}        
            \addplot[color=multicolor,mark=square,mark size=1pt,line width=2] table {
                1 57.8
                0.75 57.6
                0.5 57.5
                0.3333333333 57.6
                0.25 56.9
                0.1666666667 56.4
                0.125 54.7
                0.04166666667 51.1
                0 50.9
                };            
            \addplot[color=singlecolor,mark=square,mark size=1pt,line width=2] table {
                1 55.6
                0.75 55.7
                0.5 56.1
                0.3333333333 55.2
                0.25 53.4
                0.1666666667 50.5
                0.125 48.5
                0.04166666667 45.5
                0 45.6
            };
            \end{axis}
        \end{tikzpicture}  
    \caption{Performance on Natural Questions dev set for \lumen-Large trained on KILT vs NQ-only as a function of live proportion.}
    \label{fig:perf_vs_prop_for_nq_and_multitask}
\end{figure}
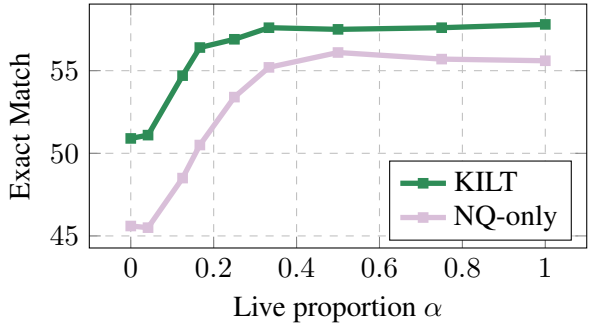
\begin{table}[ht!]
\centering
\begin{tabular}{lc}
    \textbf{Model} & \textbf{Performance} \\
    \toprule
     \model & 69.8 \\
     Frozen memory & 69.0 \\
    \bottomrule
\end{tabular}
\caption{\textbf{Training memory is a significant factor in strong \model performance.} Average performance on KILT dev sets for \model-Large with 25 retrieved and 5 selected passages, with and without training memory.}
\label{table:memory_ablation}
\end{table}

\subsection{Other ablations}

There are a number of other interesting decisions in the \model architecture and training procedure. Table \ref{table:ablations} presents ablations of some of these decisions.

The original \lumen implementation featured a separate question encoder, which was necessary because the memory encoder was not fine-tuned. Here, we update the memory encoder with multi-task training, so we opt to re-use the memory encoder for encoding the question, simplifying the architecture and reducing the number of parameters. We see that this simplification comes at a small cost in performance.

There are also a number of parameter choices regarding the reranking: the weight of the perplexity distillation loss, the temperature of the score and perplexity distributions, and the method for generating a reranking score. Over or under-weighting reranking loss leads to lower performance. However, using a lower temperature for the score and perplexity distributions does help - \citet{atlas} argue that the effect of most individual passages on perplexity is small, and a lower temperature helps distinguish those differences. Finally, it appears that using the first token of each passage performs similarly to generating a score from mean-pooled representations.

\begin{table}[ht!]
\centering
\begin{tabular}{lc}
    \textbf{Model} & \textbf{Performance} \\
    \toprule
     \model & 69.8 \\
     \midrule
     Separate Qenc & 70.0 \\
     PDist $\lambda = 0.1$ & 69.5 \\
     PDist $\lambda = 10$ & 69.5 \\
     PDist $\tau=0.1$ & 70.1  \\
     PDist $\tau = 5$ & 69.4 \\
     Mean pool & 69.8\\
    \bottomrule
\end{tabular}
\caption{\model ablations: separate question encoder, different perplexity distillation loss weight, perplexity distillation temperature, and mean pool scoring method. Each model is Large size with 25 retrievals and 5 selected passages, evaluated on the KILT dev set.}
\label{table:ablations}
\end{table}

\section{Related Work}
\label{section:related}

Retrieval augmentation~\citep{fid, retro, rag, knnlm, realm} is a powerful technique to improve language model performance by augmenting the input with additional context. Our work is focused on improving the quality-compute trade-off for retrieval-augmented language models. It does so by unifying three lines of research: late-interaction memory, late-interaction reranking, and learning to retrieve. Our approach uses the architecture skeleton from Fusion-in-Decoder~\citep{fid}, one of the most common retrieval augmented models. We employ multi-task training on KILT~\citep{kilt} as in \citet{multikilt}.

\paragraph{Memory}

Retrieval augmentation is expensive due to the additional context that needs to be processed by the language model. Memory models such as TOME~\citep{tome}, Memorizing Transformer~\citep{memorizing}, and many others~\citep{fidmemory, trime, qama, emat, memknnlm, unlimformer} attempt to avoid this cost by pre-computing representations and storing them into a memory, such that representations can be retrieved directly rather than processed on the fly. However, such approaches sacrifice quality as memory representations are not conditioned on each individual input~\citep{fidmemory, lumen}. \emph{Late-interaction memory}~\citep{lumen, lait} improves the quality of memory approaches by only partially pre-computing retrieval representations, and performing some interaction between memory and input on the fly. In particular, our work is very closely based on \lumen~\citep{lumen}.

\paragraph{Reranking}

Like the language model itself, retrieval procedures face a trade-off between expensive online ranking with full interaction~\citep{bertrank} and the more common dual encoder approaches such as DPR~\citep{dpr} and GTR~\citep{gtr} that scores based on inner product similarity with a corpus of pre-computed passage representations. 

Often different models for retrieval are applied in a pipeline approach, with an initial cheap scoring model followed by a more powerful and expensive reranker~\citep{readerguidererank, r3rerank, kgfid}. Many rerankers also make use of late interaction to obtain a good trade-off between ranking quality and speed, such as COLBERT~\citep{colbert,colbertv2}, PreTTR~\citep{prettr}, SDR~\citep{sdr}, and Poly-encoders~\citep{polyencoders}. \model combines late-interaction memory and reranking into a single model, sharing the pre-computed representations for both use cases.

\paragraph{Learning to retrieve}

Retrieval models are often trained with supervised data~\citep{dpr, gtr}, using gold retrievals from datasets such as MS-MARCO~\citep{msmarco} or TREC CAR~\citep{trec}. When selecting passage to use for retrieval-augmented generation, we have an additional signal, namely which passages are most helpful for the reader model. A number of existing works use this signal to improve retrieval~\citep{realm, emdr, raat, emdr, atlas}. We follow ATLAS~\citep{atlas} and employ perplexity distillation to train our reranker to select passages that help lower reader model perplexity.

\section{Conclusion}
\label{section:conclusion}
Retrieval-augmented language models are powerful but slow in inference, while pre-computed memory-augmented models are fast at the cost of quality. Hybrid late-interaction models such as \lumen present a good quality-compute trade-off. We introduce \model, an improved late-interaction model that also incorporates learned end-to-end reranking and multi-task training to achieve an even better trade-off. \model achieves strong gains in quality at faster speeds compared to \lumen and \fid on the KILT benchmark of knowledge-intensive tasks.

\section*{Acknowledgements}

We thank Luke Vilnis, Tania Bedrax-Weiss and others at Google Research for insightful comments and discussion.
\bibliography{custom}
\bibliographystyle{acl_natbib}

\end{document}